\documentclass[10pt,twocolumn,letterpaper]{article}

\usepackage[pagenumbers]{cvpr} 

\usepackage{graphicx}
\usepackage{amsmath}
\usepackage{amssymb}
\usepackage{booktabs}

\usepackage[pagebackref,breaklinks,colorlinks]{hyperref}

\usepackage[capitalize]{cleveref}
\crefname{section}{Sec.}{Secs.}
\Crefname{section}{Section}{Sections}
\Crefname{table}{Table}{Tables}
\crefname{table}{Tab.}{Tabs.}

\def\cvprPaperID{*****} 
\def\confName{CVPR}
\def\confYear{2023}

\begin{document}

\title{Multi-Scale Occ: 4th Place Solution for CVPR 2023 3D Occupancy Prediction Challenge}

\author{Yangyang Ding\thanks{Equal Contribution.} \quad Luying Huang\footnotemark[1]\quad Jiachen Zhong\\
SAIC AI Lab\\
{\tt\small \{dingyangyang01, huangluying, zhongjiachen\}@saicmotor.com}
}
\maketitle

\begin{abstract}
   In this report, we present the 4th place solution for CVPR 2023 3D occupancy prediction challenge. We propose a simple method called Multi-Scale Occ for occupancy prediction based on lift-splat-shoot framework, which introduces multi-scale image features for generating better multi-scale 3D voxel features with temporal fusion of multiple past frames. Post-processing including model ensemble, test-time augmentation, and class-wise thresh are adopted to further boost the final performance. As shown on the leaderboard, our proposed occupancy prediction method ranks the 4th place with 49.36 mIoU.
\end{abstract}

\section{Introduction}
\label{sec:intro}

    Recently, 3D occupancy prediction has attracted extensive attention as it is crucial for autonomous driving systems to understand geometric and semantic information in 3D scenes. 3D occupancy is more accurate and suitable for describing objects in arbitrary shape and undefined classes at a fine-grained level. 3D occupancy prediction may be performed using various modality input (e.g. LiDAR, Radar, Image). In the challenge, we focus on predicting 3D occupancy from pure multi-camera images which still remains as a challenging problem.
    
    Based on previous perception tasks including 3D object detection and semantic map segmentation, current multi-camera 3D perception methods mainly lies in two types: 1) lift-splat-shoot(LSS)-based
    \cite{huang2022bevdet4d,huang2021bevdet,philion2020lift}, lifting 2D image features to plausible 3D volume space via implicit or explicit depth estimation. 2) Transformer-based
    \cite{wei2023surroundocc,li2023voxformer,li2022bevformer}, which define 3D queries in 3D volume space and use transformer attention mechanism to query corresponding 2D image features. We build our method based on LSS-style framework, improved with long-term temporal stereo matching and multi-scale 3D features fusion to learn spatial and temporal details simultaneously. Besides, we use decoupled head to perform occupancy and semantic prediction separately in order to relieve the extreme imbalanced distribution between unoccupied voxel grids and those grids occupied by semantic classes.


\begin{figure*}
  \centering
  \includegraphics[width=0.75\linewidth]{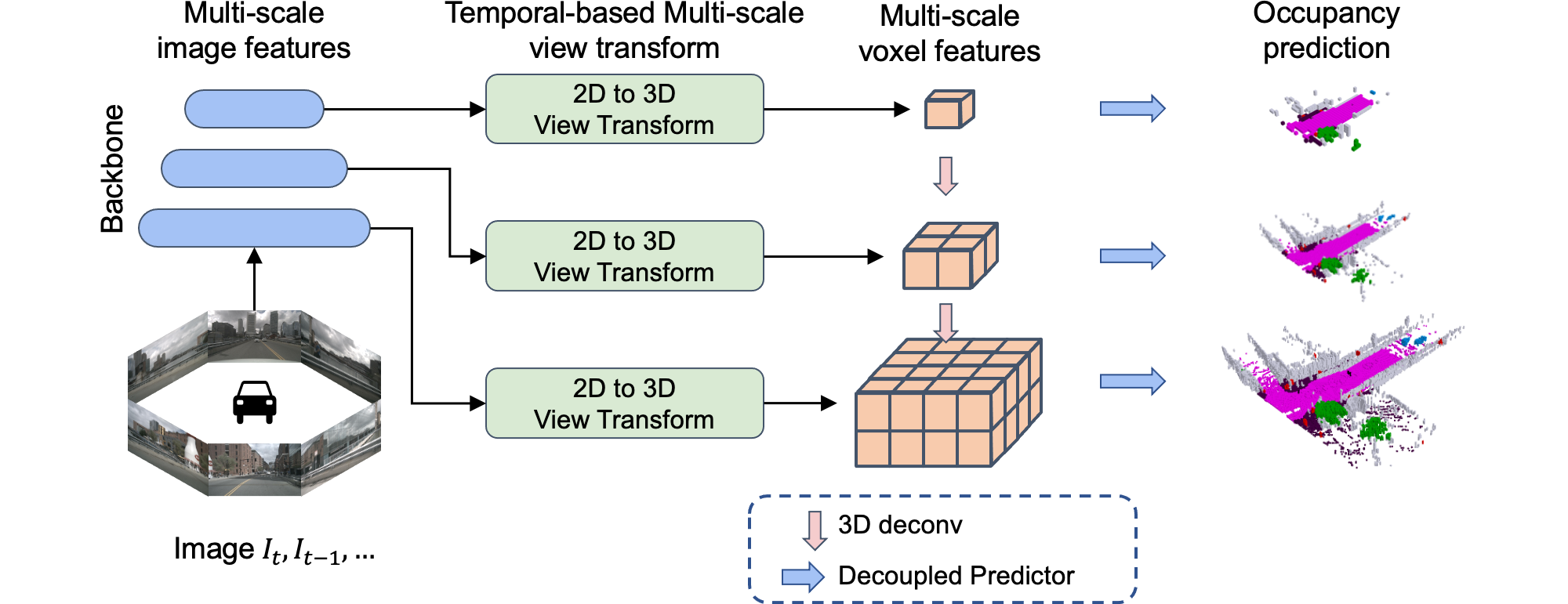}
  \caption{The architecture of our proposed Multi-scale Occ framework for 3D occupancy prediction.}
  \label{fig:short}
\end{figure*}

\section{Method}
\label{sec:2}

The overall architecture of our approach is shown in \cref{fig:short}. Given $N$ camera images with $T$ timestamps, we first use a 2D image encoder to extract $M$ scale features (\cref{sec:2.1}). Image features are then lifted to a 3D voxel feature followed by a long-term temporal feature aggregation of past frames independently on each scale to construct multi-scale 3D representation of  the current frame (\cref{sec:2.2}). To fuse multi-scale 3D features thoroughly, we use a lightweight 3D UNet~\cite{cciccek20163d} to integrate local and global geometric and semantic information (\cref{sec:2.3}). We perform occupancy and semantic prediction separately on the largest resolution with 2 decoupled heads. Multi-scale supervision is used during training to facilitate the convergence (\cref{sec:2.4}). Finally, model ensemble, test-time augmentation, and class-wise thresh are applied to further improve the performance (\cref{sec:2.5}).

\subsection{2D Backbone}
\label{sec:2.1}

For pure vision based 3D perception, 2D image backbone plays a crucial role in feature extraction from image. We use 2D backbone with a FPN~\cite{lin2017feature} to generate multi-scale 2D features. Specifically, given $N$ camera images with $T$ timestamps, the 2D backbone firstly extract three scales features for each image in strides of $\frac{1}{8}$, $\frac{1}{16}$, and $\frac{1}{32}$. Then, a simple FPN is applied to fuse features from different receptive fields and output three level features with the same channels for building 3D representation later.

In order to obtain better performance and strong generalization ability, we use Dual-InternImage-B which are composed by two InternImage-B\cite{wang2022internimage} connected via DHLC (Dense Higher-Level Composition) method\cite{liang2022cbnet} as the backbone of our main model. Moreover, we use a single Swin-B~\cite{liu2021swin} to train an extra simpler model for ensemble.

\subsection{Long-term Temporal Stereo Matching}
\label{sec:2.2}

To project multi-view and multi-timestamp 2D features to 3D space, we apply LSS view transformer~\cite{philion2020lift} with BEVPoolv2~\cite{huang2022bevpoolv2} to create 3D voxel features at each scale. In LSS-based mechanism, accurate depth estimation is important to performance~\cite{li2022bevdepth} and temporal stereo style methods are very helpful in improving model depth prediction ability~\cite{yao2018mvsnet, li2022bevstereo, Park2022TimeWT}. Therefore, following~\cite{Park2022TimeWT}, we use
the total of T = 9 timestamps(1 current frame + 8 past frames) to create cost volumes to enhance depth estimation. Specifically, we create total of 8 frames cost volumes using image features at the stride of $\frac{1}{4}$ between each two adjacent frames. Then those cost volumes are rescaled to the strides of $\frac{1}{8}$, $\frac{1}{16}$, and $\frac{1}{32}$ and concatenated with image features at each corresponding scale to predict the depth. Notice that, the earliest frame is dropped after building cost volumes, thus, 8 frames of 3D voxel features are lifted in total. We use ego-motion to align previous frames to current frame following BEVDet4D~\cite{huang2022bevdet4d} and concatenate them along channel dimension. Finally, we feed the concatenated temporal aligned feature to ResNet and LSS-FPN following~\cite{huang2021bevdet} but replacing the 2D convolution with 3D to further fuse temporal information. Notice that, as shown in~\ref{fig:short}, we project 3 scales of image features independently to construct 3 different scales of 3D voxel features. Besides depth cost volume are constructed at $\frac{1}{4}$ stride, the temporal fusion happens also individually in different scales of 3D voxel features.


\subsection{Multi-scale 3D Features Fusion }
\label{sec:2.3}
In computer vision tasks, multi-scale features fusion and supervision often achieve better results~\cite{ronneberger2015unet, lin2017feature}. Recent work~\cite{wei2023surroundocc} indicates that constructing 3D representation at multi-scale independently improves performance in occupancy prediction. Inspired by the idea, as described in \cref{sec:2.2}, we construct 3 different scales of 3D voxel features ($50\times50\times4$, $100\times100\times8$, and $200\times200\times16$) individually from corresponding scales of 2D image features ($\frac{1}{32}$, $\frac{1}{16}$, $\frac{1}{8}$). Different from~\cite{wei2023surroundocc} which does not utilize temporal information, we perform extra long-term temporal fusion individually at each scale of 3D voxel features. After individually building 3 scales of 3D voxel features, we use a 3D U-Net~\cite{cciccek20163d} to further fuse multi-scale 3D features. Supervision is applied at each scale.

\subsection{Decoupled Prediction Head}
\label{sec:2.4}
Class-imbalanced problem naturally exists in real-world vision tasks which also occurs in this challenge. The class of free, which indicates the occupancy status of voxel grid,  occurs much more frequent than other semantic classes (free class takes around $96\%$ in training set). To tackle this problem, we decouple the occupancy
prediction head and semantic prediction head. Specifically, given the 3D feature volume $V_{i}$ generated at each scale $i=0,1,2$ , two parallel prediction heads with different output channels are applied to predict whether each voxel is occupied or not (1 class), and which semantic label (16 classes) it belongs to. Each head is a simple 2-layer MLP with Softplus activation.


For occupancy prediction head, we use binary cross-entropy loss $L_{i}^{occ}$. For semantic prediction head, we adopt multi-class focal loss $L_{i}^{sem}$~\cite{lin2017focal}. Besides, we only compute the loss of those voxel grids which can be observed in the current camera view by using the binary voxel mask $mask_{i}^{cam}$ during training:
\begin{equation}\label{eq3}
L_{i}^{occ}=BCE(V_{i}, GT_{i}^{occ}) * mask_{i}^{cam},
\end{equation}
\begin{equation}\label{eq3}
L_{i}^{sem}=FL(V_{i}, GT_{i}^{sem}) * mask_{i}^{cam},
\end{equation}
where $GT_{i}^{occ}$ is the geometric ground truth with $\{0, 1\}$ label (0 for unoccupied and 1 for occupied). $GT_{i}^{sem}$ is the semantic ground truth with 16 classes. A lower resolution of $GT_{i}^{sem}$ is obtained from majority vote pooling~\cite{roldão2020lmscnet} of the full size ground truth and $GT_{i}^{occ}$ is obtained via max pooling. For better handling data imbalance, we also re-weight each class loss with the inverse of the class-frequency as in~\cite{roldão2020lmscnet}, which is applied to both occupancy head and semantic head individually. Finally,  our model is trained by minimizing the following objective:
\begin{equation}\label{eq3}
L_{i}=L_{i}^{occ}+L_{i}^{sem}+L_{i}^{depth},
\end{equation}
\begin{equation}\label{eq3}
L_{total}=\sum_{i=0}^{2}\alpha_iL_{i}, 
\end{equation}
We rescale the loss on $i$-th scale via $\alpha_i=\frac{1}{2^i}$ to make sure larger resolution prediction plays more important role in training.

\subsection{Post-process}
\label{sec:2.5}

In order to further enhance the performance of our model, we adopt model ensemble and test-time augmentation. We train one extra simpler model with single Swin-B\cite{liu2021swin} as backbone without multi-scale supervision. We apply test-time augmentation separately to both models. We use image horizontal flip, 3D space horizontal and vertical flips as augmentation. Each model inferences 8 times, so that we can obtain 16 occupancy predictions and 16 semantic predictions for each frame. Due to performance difference between the two models, we weight Dual-InternImage-B model with coefficient 0.55 and Swin-B model with 0.45. We fuse the prediction results to obtain final results $P_{occ}$ and $P_{sem}$ using the following formulation:

\begin{equation}\label{eq1}
P_{occ}=0.45\sum_{i=1}^{16} P_{occ1_i}+0.55\sum_{i=1}^{16} P_{occ2_i}
\end{equation}
\begin{equation}\label{eq2}
P_{sem}=argmax(0.45\sum_{i=1}^{16} P_{sem1_i}+0.55\sum_{i=1}^{16} P_{sem2_i})
\end{equation}
where \(P_{occ1_i}\) and \(P_{sem1_i}\) denote as the $i$-th test-time augmentation prediction probability of the Swin-B model, \(P_{occ2_i}\) and \(P_{sem2_i}\) stand for the prediction of Dual-InternImage-B model.

To obtain the ultimate prediction, we select threshold for each category. If the occupancy prediction for a voxel grid is below the threshold, it is considered as unoccupied. The threshold for each category are shown in Table \ref{tab:1}.

\begin{table}
  \centering
  \begin{tabular}{@{}lc@{}}
    \toprule
    Class & Threshold \\
    \midrule
    Others & 0.92 \\
    Barrier & 0.94 \\
    Bicycle & 0.94 \\
    Bus & 0.94 \\
    Car & 0.93 \\
    Construction Vehicle & 0.93 \\
    Motorcycle & 0.91 \\
    Pedestrian & 0.91 \\
    Traffic Cone & 0.91 \\
    Trailer & 0.93 \\
    Truck & 0.93 \\
    Driveable Surface & 0.96 \\
    Other Flat & 0.95 \\
    Sidewalk & 0.95 \\
    Terrain & 0.95 \\
    Manmade & 0.93 \\
    Vegetation & 0.92\\
    \bottomrule
  \end{tabular}
  \caption{Threshold for each class during post-process.}
  \label{tab:1}
\end{table}

\section{Experiments}

\subsection{Experimental Setup}

\textbf{Dataset.} The challenge dataset contains 28130 frames for training, 6019 for validation, and 6008 for testing respectively. Each frame contains 6 views of camera images with $1600\times900$ resolution. 


\textbf{Architecture. } We initialize the two InternImage-B backbones with the same weight from the official repository~\cite{wang2022internimage} which is trained for COCO object detection and instance segmentation task via Mask-RCNN~\cite{he2018mask} method. Swin-B backbone is initialized with the weight from BEVDet4D~\cite{huang2022bevdet4d} official repository which is trained for nuScenes 3D object detection task. The input image resolution is resized to $512\times1408$ during training and inference. 

\textbf{Training Details.} We use AdamW optimizer~\cite{loshchilov2017decoupled} with a constant learning rate 2e-4 through training and apply Exponential Moving Average(EMA) strategy with average factor 0.999 to update our model. Our Swin-B model are trained for 24 epochs with weight decay 0.01 on 14 Tesla V100 GPUs. And our Dual-InternImage-B model are totally trained for 31 epochs, in which the first 6 epochs is trained with weight decay 0.05 on 24 Tesla V100 GPUs and the last 25 epochs is trained with weight decay 0.01 on 31 Tesla V100 GPUs. We keep batch size on single GPU equal to $2$ for all training. Both two models are trained on training and validation set for leaderboard submission without training hyperparameters tuning. 

\subsection{Main Results}
Our three submission results are presented in~\ref{tab:2}, each submission improves  the performance compared with previous one. Due to the limitation of computation resources, we are unable to perform comprehensive ablation study of different settings. We may only roughly summarize that stronger image backbone, multi-scale 3D voxel features, model ensemble, and test-time augmentation contribute positively to the performance of this task. Our best submission ranks the 4th place with 49.36 mIoU on the leaderboard.
\begin{table} 
  \centering
  \begin{tabular}{@{}lc@{}}
    \toprule
    Method & mIoU \\
    \midrule
    Swin-B + Prev8 & 46.55 \\
    Dual-InternImage-B + Prev8 + Multi-Scale & 48.00 \\
    Model Ensemble + TTA & 49.36 \\
    \bottomrule
  \end{tabular}
  \caption{Submission results on the test set.}
  \label{tab:2}
\end{table}

\section{Conclusion}
In this technical report, we present our Multi-Scale Occ solution for 3D occupancy prediction. In order to achieve better prediction results, we construct multi-scale 3D voxel features with long-term temporal information fusion individually at each scale. 3D U-Net and decoupled heads are applied to perform fusion and prediction of multiple scales. In addition, we conduct model ensemble, test-time augmentation, and class-wise thresh selection to enhance the performance. Our overall 3D occupancy prediction framework achieves the 4th place in the CVPR 2023 3D occupancy prediction challenge.

\section*{Acknowledgements}
We would like to thank our colleagues Yineng Xiong and Zeyuan Zhang for GPU resources coordination.

{\small
\bibliographystyle{ieee_fullname}
\bibliography{egbib.bib}

\begin{thebibliography}{10}\itemsep=-1pt

\bibitem{cciccek20163d}
{\"O}zg{\"u}n {\c{C}}i{\c{c}}ek, Ahmed Abdulkadir, Soeren~S Lienkamp, Thomas
  Brox, and Olaf Ronneberger.
\newblock 3d u-net: learning dense volumetric segmentation from sparse
  annotation.
\newblock In {\em Medical Image Computing and Computer-Assisted
  Intervention--MICCAI 2016: 19th International Conference, Athens, Greece,
  October 17-21, 2016, Proceedings, Part II 19}, pages 424--432. Springer,
  2016.

\bibitem{he2018mask}
Kaiming He, Georgia Gkioxari, Piotr Dollár, and Ross Girshick.
\newblock Mask r-cnn, 2018.

\bibitem{huang2022bevdet4d}
Junjie Huang and Guan Huang.
\newblock Bevdet4d: Exploit temporal cues in multi-camera 3d object detection.
\newblock {\em arXiv preprint arXiv:2203.17054}, 2022.

\bibitem{huang2022bevpoolv2}
Junjie Huang and Guan Huang.
\newblock Bevpoolv2: A cutting-edge implementation of bevdet toward deployment.
\newblock {\em arXiv preprint arXiv:2211.17111}, 2022.

\bibitem{huang2021bevdet}
Junjie Huang, Guan Huang, Zheng Zhu, Ye Yun, and Dalong Du.
\newblock Bevdet: High-performance multi-camera 3d object detection in
  bird-eye-view.
\newblock {\em arXiv preprint arXiv:2112.11790}, 2021.

\bibitem{li2022bevstereo}
Yinhao Li, Han Bao, Zheng Ge, Jinrong Yang, Jianjian Sun, and Zeming Li.
\newblock Bevstereo: Enhancing depth estimation in multi-view 3d object
  detection with dynamic temporal stereo, 2022.

\bibitem{li2022bevdepth}
Yinhao Li, Zheng Ge, Guanyi Yu, Jinrong Yang, Zengran Wang, Yukang Shi,
  Jianjian Sun, and Zeming Li.
\newblock Bevdepth: Acquisition of reliable depth for multi-view 3d object
  detection.
\newblock {\em arXiv preprint arXiv:2206.10092}, 2022.

\bibitem{li2023voxformer}
Yiming Li, Zhiding Yu, Christopher Choy, Chaowei Xiao, Jose~M Alvarez, Sanja
  Fidler, Chen Feng, and Anima Anandkumar.
\newblock Voxformer: Sparse voxel transformer for camera-based 3d semantic
  scene completion.
\newblock In {\em Proceedings of the IEEE/CVF Conference on Computer Vision and
  Pattern Recognition (CVPR)}, 2023.

\bibitem{li2022bevformer}
Zhiqi Li, Wenhai Wang, Hongyang Li, Enze Xie, Chonghao Sima, Tong Lu, Yu Qiao,
  and Jifeng Dai.
\newblock Bevformer: Learning bird’s-eye-view representation from
  multi-camera images via spatiotemporal transformers.
\newblock {\em arXiv preprint arXiv:2203.17270}, 2022.

\bibitem{liang2022cbnet}
Tingting Liang, Xiaojie Chu, Yudong Liu, Yongtao Wang, Zhi Tang, Wei Chu,
  Jingdong Chen, and Haibin Ling.
\newblock Cbnet: A composite backbone network architecture for object
  detection.
\newblock {\em IEEE Transactions on Image Processing}, 31:6893--6906, 2022.

\bibitem{lin2017feature}
Tsung-Yi Lin, Piotr Doll{\'a}r, Ross Girshick, Kaiming He, Bharath Hariharan,
  and Serge Belongie.
\newblock Feature pyramid networks for object detection.
\newblock In {\em Proceedings of the IEEE conference on computer vision and
  pattern recognition}, pages 2117--2125, 2017.

\bibitem{lin2017focal}
Tsung-Yi Lin, Priya Goyal, Ross Girshick, Kaiming He, and Piotr Doll{\'a}r.
\newblock Focal loss for dense object detection.
\newblock In {\em Proceedings of the IEEE international conference on computer
  vision}, pages 2980--2988, 2017.

\bibitem{liu2021swin}
Ze Liu, Yutong Lin, Yue Cao, Han Hu, Yixuan Wei, Zheng Zhang, Stephen Lin, and
  Baining Guo.
\newblock Swin transformer: Hierarchical vision transformer using shifted
  windows, 2021.

\bibitem{loshchilov2017decoupled}
Ilya Loshchilov and Frank Hutter.
\newblock Decoupled weight decay regularization.
\newblock {\em arXiv preprint arXiv:1711.05101}, 2017.

\bibitem{Park2022TimeWT}
Jinhyung Park, Chenfeng Xu, Shijia Yang, Kurt Keutzer, Kris Kitani, Masayoshi
  Tomizuka, and Wei Zhan.
\newblock Time will tell: New outlooks and a baseline for temporal multi-view
  3d object detection.
\newblock 2023.

\bibitem{philion2020lift}
Jonah Philion and Sanja Fidler.
\newblock Lift, splat, shoot: Encoding images from arbitrary camera rigs by
  implicitly unprojecting to 3d.
\newblock In {\em Proceedings of the European Conference on Computer Vision},
  2020.

\bibitem{roldão2020lmscnet}
Luis Roldão, Raoul de Charette, and Anne Verroust-Blondet.
\newblock Lmscnet: Lightweight multiscale 3d semantic completion, 2020.

\bibitem{ronneberger2015unet}
Olaf Ronneberger, Philipp Fischer, and Thomas Brox.
\newblock U-net: Convolutional networks for biomedical image segmentation,
  2015.

\bibitem{wang2022internimage}
Wenhai Wang, Jifeng Dai, Zhe Chen, Zhenhang Huang, Zhiqi Li, Xizhou Zhu,
  Xiaowei Hu, Tong Lu, Lewei Lu, Hongsheng Li, et~al.
\newblock Internimage: Exploring large-scale vision foundation models with
  deformable convolutions.
\newblock {\em arXiv preprint arXiv:2211.05778}, 2022.

\bibitem{wei2023surroundocc}
Yi Wei, Linqing Zhao, Wenzhao Zheng, Zheng Zhu, Jie Zhou, and Jiwen Lu.
\newblock Surroundocc: Multi-camera 3d occupancy prediction for autonomous
  driving.
\newblock {\em arXiv preprint arXiv:2303.09551}, 2023.

\bibitem{yao2018mvsnet}
Yao Yao, Zixin Luo, Shiwei Li, Tian Fang, and Long Quan.
\newblock Mvsnet: Depth inference for unstructured multi-view stereo.
\newblock In {\em Proceedings of the European conference on computer vision
  (ECCV)}, pages 767--783, 2018.

\end{thebibliography}
}

\end{document}